\documentclass{article}
\usepackage{float}
\usepackage{graphicx} 
\usepackage{amsmath,amssymb}
\usepackage{booktabs}
\usepackage{algorithm}
\usepackage{algpseudocode}
\usepackage[authoryear,round]{natbib}
\usepackage[hidelinks]{hyperref}

\title{G-NAC: Graph Neural Automata Clustering via Emergent Domain Formation}
\author{Keith Miller, Tristan Crawford, Ronald J. Stanley, PhD}
\date{September 2026}

\begin{document}

\maketitle

\begin{abstract}
We introduce Graph Neural Automata Clustering (G-NAC), an unsupervised
method for partitioning numerical feature vectors without cluster-label
supervision. Each observation is represented as a cell on a fixed
neighborhood graph, and a shared recurrent graph-neural cellular rule
evolves hidden and domain states through local interactions. The final
output is one cluster assignment per observation; reference partitions
are withheld from model fitting and used only for external evaluation.
G-NAC combines graph smoothness with variance, covariance, dispersion,
and temporal regularization, then converts the evolved relational state
into a rank-based sparse spectral affinity for partitioning. We evaluate
the complete pipeline on 73 dataset/$K$ tasks from 57 datasets in the
four established clustering benchmark batteries. G-NAC achieved
mean ARI 0.7951, comparable in observed mean to Genie (0.7941) and higher
than the other evaluated baselines. Controlled corruption experiments
showed little aggregate degradation when up to half of each node's
directed neighbor slots were deleted and retraining was performed, but
progressive degradation under edge rewiring and measurement noise. For
fixed graph degree and model configuration, empirical fit time and peak
allocated GPU memory scaled approximately as $N^{1.02}$ and $N^{0.99}$,
respectively, over $5{,}000$--$100{,}000$ nodes. Finally, frozen
transition rules trained on smaller source graphs transferred to
independent 100,000-node samples from the same generating process; on the
harder noisy-blobs family, training on 5,000 source observations reduced
target ARI by less than 0.009 relative to 100,000-node source training.
These results support G-NAC as a recurrent graph-clustering formulation
while identifying dependencies on graph quality, readout design, and
matched source--target structure.
\end{abstract}

\section{Introduction}

Clustering seeks to recover meaningful structure from unlabeled
observations, where each observation is represented by a numerical
feature vector $x_i\in\mathbb{R}^d$. Classical approaches include
$k$-means \citep{MacQueen1967KMeans}, Gaussian mixture modeling
commonly fit by expectation--maximization \citep{Dempster1977EM},
hierarchical methods such as Ward clustering \citep{Ward1963}, and
graph-based spectral clustering \citep{vonLuxburg2007}. Modern deep
clustering methods instead learn representations intended to facilitate
partitioning. Despite substantial differences in implementation, these
approaches generally seek either a partition of the observations or a
representation from which such a partition can be obtained.

In the unsupervised setting considered here, the objective is to assign
the observations to $K$ groups without using reference cluster labels
during model fitting. Where benchmark reference partitions are
available, they are used only after clustering to quantify agreement
with the recovered partition rather than as targets for training.

An alternative is to treat clustering as an emergent dynamical process
in which repeated local interactions between observations reorganize an
initially ambiguous system into coherent domains. This perspective has
a long history in cellular automata and self-organizing systems,
including cellular-automata and cellular-learning-automata approaches
to clustering and community detection
\citep{Shuai2007GeneralizedCA,deLopeMaravall2013,
DundarKorkmaz2018,Esnaashari2007ICLA,Zhao2015CLANet}. These methods,
however, generally employ prescribed stochastic or learning-automata
rules rather than a differentiable graph-neural cellular transition rule
learned directly from the unlabeled observations.

Graph Neural Cellular Automata (GNCA) provide a natural framework for
learning such dynamics \citep{Grattarola2021}. A GNCA associates states
with graph nodes and repeatedly applies a shared neural transition
function through local graph interactions. This raises a different
possibility for unsupervised learning: rather than training a neural
network to predict clusters, can a learned cellular system organize
observations into clusters through its recurrent dynamics?

We investigate this question through Graph Neural Automata Clustering
(G-NAC). A fixed neighborhood graph specifies which observations may
interact, while a shared learned transition rule recurrently evolves
hidden and domain states at every node. Training balances local graph
coherence with variance, covariance, and dispersion regularizers that
penalize representational collapse. Rather than directly predicting
cluster labels, G-NAC learns a common rule of local evolution through
which coherent latent domains can emerge. To the best
of our knowledge, we have not identified prior work that trains a GNCA
transition rule itself as a general-purpose unsupervised data-clustering
mechanism. This novelty claim is intentionally narrow: cellular automata
have previously been used for clustering, and graph neural networks have
extensively been used for graph clustering.

Useful G-NAC organization need not correspond to coordinate convergence
toward a fixed state. Instead, the relative ordering of domain-space
distances along graph edges can exhibit high agreement across successive
recurrent checkpoints while the coordinates themselves continue to
change. G-NAC therefore uses successive edge-distance rank agreement as
an empirical stopping criterion and constructs its final sparse spectral
affinity from global ordinal ranks of domain-space edge distances. The
rank criterion measures relational stability rather than certifying
fixed-point convergence.

\begin{figure*}[t]
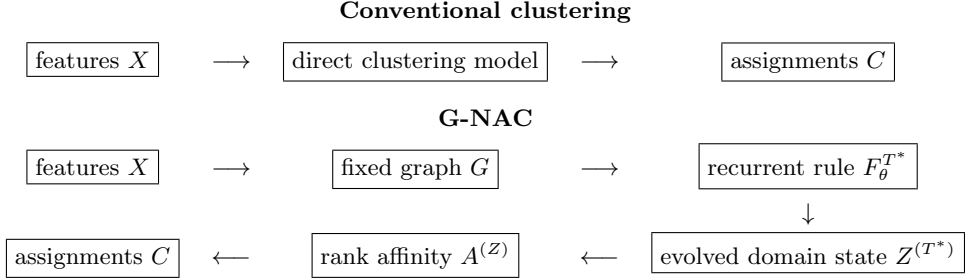

    \centering
    \small
    \renewcommand{\arraystretch}{1.7}
    \begin{tabular}{ccccc}
    \multicolumn{5}{c}{\textbf{Conventional clustering}}\\
    \fbox{features $X$}
    & $\longrightarrow$ &
    \fbox{direct clustering model}
    & $\longrightarrow$ &
    \fbox{assignments $C$}\\[3pt]
    \multicolumn{5}{c}{\textbf{G-NAC}}\\
    \fbox{features $X$}
    & $\longrightarrow$ &
    \fbox{fixed graph $G$}
    & $\longrightarrow$ &
    \fbox{recurrent rule $F_\theta^{T^*}$}\\[-2pt]
    & & & & $\downarrow$\\[-2pt]
    \fbox{assignments $C$}
    & $\longleftarrow$ &
    \fbox{rank affinity $A^{(Z)}$}
    & $\longleftarrow$ &
    \fbox{evolved domain state $Z^{(T^*)}$}
    \end{tabular}
    \caption{Conceptual comparison of direct clustering and G-NAC. G-NAC
    inserts a learned recurrent graph-neural cellular process between the
    observed feature vectors and the final partition. Reference labels are
    not inputs to the recurrent model.}
    \label{fig:conceptual_comparison}
\end{figure*}

Figure~\ref{fig:conceptual_comparison} summarizes the distinction between
a conventional direct clustering pipeline and G-NAC. The complete state
updates, objective, stopping rule, and readout equations are given in
Section~\ref{sec:methods}.

We evaluate G-NAC on 73 dataset/$K$ tasks drawn from 57 externally
curated clustering datasets against classical, spectral,
self-organizing, and deep clustering baselines. We further examine
stochastic stability, measurement and graph corruption, computational
scaling, and transfer of learned cellular rules from smaller source
graphs to independent 100,000-node target graphs. Section~\ref{sec:data_tasks}
provides the benchmark composition and clarifies the experimental task
and evaluation protocol.

The principal contributions of this work are:

\begin{itemize}
    \item We introduce G-NAC, an unsupervised clustering framework in
    which a shared graph-neural cellular transition rule organizes
    observations into latent domains through recurrent local
    interaction.

    \item We formulate an unsupervised domain-formation objective that
    combines local graph coherence with variance, covariance, and
    dispersion regularization to encourage coherent, noncollapsed
    representations without cluster-label supervision.

    \item We introduce a relational inference procedure based on
    graph-edge distance-rank stability and a global edge-rank affinity.
    At fixed graph support and tie convention, the resulting affinity
    is invariant to strictly increasing transformations of the learned
    edge distances.

    \item We evaluate G-NAC against ten classical, spectral,
    self-organizing, and deep clustering baselines and characterize its
    stochastic stability, response to measurement and graph
    corruption, and empirical computational scaling.

    \item We demonstrate transfer of learned cellular rules across
    graph scale and independent graph realizations under matched
    generating conditions, including deployment of rules trained on
    smaller source graphs to independent 100,000-node targets without
    retraining.
\end{itemize}

These results establish learned domain formation as an alternative
formulation of clustering in which a shared local dynamical rule
reorganizes a relational system before partitioning.


\section{Related Work}
\label{sec:related_work}

G-NAC lies at the intersection of several established research
directions: cellular-automata-based clustering, graph neural cellular
automata, deep representation learning for clustering, and
graph-structured clustering. These areas provide important precedents
for individual components of the proposed method, but differ from
G-NAC in the role assigned to the recurrent cellular dynamics.

\subsection{Cellular Automata for Data Clustering}
\label{sec:ca_clustering}

Cellular automata have been used for unsupervised clustering well
before the development of modern neural cellular automata. Early
approaches treated observations as cells whose local interactions
reorganize the data into groups. Generalized cellular automata
formulated clustering as a stochastic self-organizing process
\citep{Shuai2007GeneralizedCA}, while \citet{deLopeMaravall2013}
proposed a linear cellular-automata clustering algorithm in which
observations are reorganized through local dynamics. Stochastic
cellular automata have likewise been applied directly to data
clustering \citep{DundarKorkmaz2018}. These methods demonstrate that
local cellular interactions can produce unsupervised partitions, but
use algorithmically prescribed rather than learned neural transition
rules.

Related work in cellular learning automata (CLA) extends cellular
organization to irregular graph structures. \citet{Esnaashari2007ICLA}
introduced irregular cellular learning automata (ICLA), allowing cells
to interact over arbitrary graph topologies, and applied the framework
to clustering in sensor networks. Building on ICLA,
\citet{Zhao2015CLANet} proposed CLA-net for community detection, in
which graph nodes contain learning automata whose collective evolution
reveals community structure.

These approaches are structurally related to G-NAC but employ a
different learning mechanism. Cellular learning automata update
action-probability distributions according to learning-automata rules,
whereas G-NAC shares a differentiable neural transition function across
cells and trains it through recurrent gradient-based optimization.
Moreover, CLA-based community-detection methods primarily seek
structure represented by the input graph, while G-NAC uses a
neighborhood graph as the interaction substrate over which learned
latent domain states evolve.

\subsection{Graph Neural Cellular Automata}
\label{sec:gnca_related}

Neural cellular automata replace hand-designed transition rules with
learned neural functions, and Graph Neural Cellular Automata (GNCA)
generalize this formulation from regular lattices to arbitrary graph
topologies. \citet{Grattarola2021} introduced a framework in which a
graph neural network parameterizes a shared cellular transition rule,
demonstrating that repeated local graph interactions can produce
coordinated global dynamics across tasks including pattern formation
and dynamical-system simulation.

This formulation provides the principal architectural foundation for
G-NAC. Both approaches associate recurrent states with graph vertices,
reuse a shared neural transition function over time, and obtain global
organization through repeated local message passing. They differ
primarily in objective and interpretation: the original GNCA framework
learns dynamics for specified tasks or target configurations, whereas
G-NAC uses an unsupervised domain-formation objective and interprets
the resulting relational organization as cluster structure.

Subsequent work has extended GNCA to learned graph topology
\citep{DwyerOmwenga2023}, E$(n)$-equivariant pattern formation and
dynamical modeling \citep{Gala2024EGNCA}, and physics-informed
dynamical systems \citep{Navarin2024PhysicsGNCA}. These extensions
broaden the GNCA framework but do not formulate the learned cellular
rule as a general-purpose unsupervised data-clustering mechanism.

The novelty boundary is therefore intentionally narrow. Cellular
automata have previously been used for clustering, and graph neural
networks have extensively been used for graph clustering; G-NAC instead
uses a \emph{shared recurrent graph-neural cellular rule} whose
unsupervised dynamics generate a cluster-revealing state.

\subsection{Deep Representation Learning for Clustering}
\label{sec:deep_clustering}

A large body of deep clustering work instead approaches clustering as
a representation-learning problem. Deep Embedded Clustering (DEC)
jointly learns a nonlinear mapping into a latent space and refines
cluster assignments using a clustering-oriented objective
\citep{Xie2016DEC}. Improved Deep Embedded Clustering (IDEC) augments
this formulation with reconstruction-based structure preservation,
reducing distortion of the learned feature space during clustering
\citep{Guo2017IDEC}.

These methods differ fundamentally from G-NAC in their computational
organization. DEC and IDEC learn a feed-forward embedding in which
observations become easier to partition. G-NAC instead learns a
recurrently applied local transition rule. The representation of a
cell therefore depends not only on its initial feature vector but also
on the evolving states of its graph neighbors over multiple recurrent
steps.

This distinction is relevant because G-NAC does not optimize
point-wise cluster assignments directly. Rather, clustering emerges
from repeated local interactions constrained by a graph substrate.
However, DEC and IDEC are important comparison methods because
they represent the established paradigm of explicitly learning a latent
representation for unsupervised clustering.

The unsupervised objective of G-NAC also draws on recent
self-supervised representation-learning methods to prevent
representational collapse. Variance and covariance regularization
are inspired by VICReg \citep{Bardes2022}, which maintains
feature-wise variance while discouraging redundant latent dimensions.
The term non-neighbor dispersion is related to the objective of Gaussian-potential
uniformity of \citet{WangIsola2020}. These methods were
primarily developed for representation learning over independently
sampled observations. G-NAC adapts related anti-collapse principles
to an evolving graph-connected domain state, balancing local
graph smoothness against global representational dispersion.

\subsection{Deep Graph Clustering}
\label{sec:deep_graph_clustering}

Deep graph clustering methods combine graph representation learning
with unsupervised clustering objectives. Representative approaches
include DAEGC, which couples graph-attention embeddings with
self-training \citep{Wang2019DAEGC}, and SDCN, which combines
autoencoding with graph convolution to incorporate structural
information during clustering \citep{Bo2020SDCN}. DMoN instead learns
graph representations through a differentiable modularity objective,
directly optimizing graph partition structure without pseudo-label
supervision \citep{Tsitsulin2023DMoN}.

G-NAC likewise exploits local graph structure, but differs in both its
dynamics and the role assigned to the graph. Rather than using a
finite feed-forward graph encoder, G-NAC repeatedly applies a shared
cellular transition rule, making the learned object a rule of graph-state
evolution rather than only an embedding function. Moreover, its $k$NN
graph is constructed from observations and serves as an interaction
substrate rather than necessarily being the object to be partitioned.
The recurrent dynamics instead transform latent relations on this
fixed topology before the resulting representation is clustered.

\subsection{Self-Organizing Neural Clustering}
\label{sec:self_organizing}

G-NAC is also conceptually related to self-organizing neural
clustering methods. The Self-Organizing Map (SOM)
\citep{Kohonen1982SOM}, Neural Gas \citep{Martinetz1991NeuralGas},
and Growing Neural Gas \citep{Fritzke1995GNG} organize
representative units according to the distribution and topology of
observed data. Such methods demonstrate that useful cluster structure
can emerge through repeated local or competitive adaptation rather
than explicit supervised targets.

The similarity is primarily conceptual. In both self-organizing
clustering and G-NAC, global organization emerges from repeated local
adaptation. However, classical competitive-learning methods update
prototype vectors or network topology according to competitive
adaptation rules. G-NAC instead maintains one graph cell per
observation and learns a shared neural transition rule that updates
latent cellular states through conductance-weighted neighbor
interactions. The graph dynamics therefore operate directly over the
observations rather than through a separate set of learned prototypes.

\subsection{Relational and Rank-Based Cluster Readout}
\label{sec:rank_related}

The final G-NAC readout is related to rank-based similarity and
spectral clustering. Point-wise neighbor orderings have been used as
similarity signals in rank-order clustering \citep{Zhu2011RankOrder},
while shared-neighbor information has been used to construct
affinities for spectral clustering \citep{YeSakurai2016}. G-NAC
instead globally ranks learned domain-space distances over the edges
of the fixed graph. The same relational ordering underlies both the
empirical rank-stability stopping criterion and the final affinity
construction, linking recurrent evolution to the partition readout.

Learned spectral clustering provides a complementary precedent.
SpectralNet learns a nonlinear representation by approximately
optimizing a spectral clustering objective \citep{Shaham2018SpectralNet}.
G-NAC instead learns a shared recurrent cellular transition rule
through local graph interactions and applies spectral partitioning
only to the resulting edge-rank affinity. Spectral clustering
therefore serves as a readout of the learned cellular organization
rather than as the objective defining the representation.

\subsection{Positioning of G-NAC}
\label{sec:positioning}

The preceding literature establishes several important precedents.
Cellular automata have been used directly for data clustering;
cellular learning automata have been used for community detection;
GNCAs provide learned recurrent dynamics on arbitrary graphs; and deep
graph clustering learns graph-informed representations for
unsupervised partitioning. G-NAC combines these themes but assigns a
different role to the cellular dynamics.

Specifically, G-NAC learns a shared graph-neural cellular transition
rule whose recurrent unsupervised evolution transforms a fixed
neighborhood graph into a cluster-revealing relational state. Cluster
structure is therefore treated as an emergent domain organization of
the learned graph dynamics rather than as the direct output of a
feed-forward clustering head, a hand-designed cellular rule, or a
community-detection objective applied to the original graph.

\section{Method}
\label{sec:methods}

\subsection{Problem Formulation}
\label{sec:problem}

Let
\begin{equation}
    X = \{x_i\}_{i=1}^{N},
    \qquad
    x_i \in \mathbb{R}^{d},
\end{equation}
denote $N$ observations represented by $d$-dimensional feature
vectors. Graph Neural Automata Clustering (G-NAC) seeks a partition
\begin{equation}
    C = \{c_i\}_{i=1}^{N},
    \qquad
    c_i \in \{1,\ldots,K\},
\end{equation}
without cluster-label supervision. Rather than predicting assignments
directly, G-NAC represents observations as cells on a graph and learns
a shared local transition rule whose repeated application produces a
cluster-revealing relational state:
\begin{equation}
    X,G,Z^{(0)}
    \longrightarrow
    Z^{(1)}
    \longrightarrow
    \cdots
    \longrightarrow
    Z^{(T)}
    \longrightarrow C.
\end{equation}

The requested number of clusters $K$ is used only by the final
component-resolved or spectral readout; it is not supplied during
training or recurrent inference. In the benchmark experiments, $K$ is
taken from the reference-partition metadata and supplied equally to
methods requiring a cluster count; in deployment it must be specified
externally or estimated by a separate procedure. Input preprocessing is
likewise external to G-NAC, and the estimator does not internally
standardize $X$.

The formulation requires $N>1$, $d_z>1$, at least one graph edge, and
graph non-neighbor pairs for the dispersion term. Small positive
offsets are used in local-scale and bandwidth calculations to handle
zero distances numerically; inputs violating the structural
requirements require explicit handling.

\subsection{Fixed Neighborhood Graph and Edge Geometry}
\label{sec:graph}

G-NAC first constructs a fixed Euclidean $k$-nearest-neighbor (kNN)
graph
\begin{equation}
    G=(V,E),
\end{equation}
with one graph vertex for each observation. For every vertex, the $k$
nearest neighbors are computed using Euclidean distance. Because the
directed kNN relation is not generally symmetric, it is converted to
an undirected relation by union:
\begin{equation}
    (i,j)\in E
    \iff
    j\in\mathcal{N}_{k}(i)
    \;\lor\;
    i\in\mathcal{N}_{k}(j).
\end{equation}
This is the conventional symmetrized kNN construction used in
similarity-graph methods and spectral clustering
\citep{vonLuxburg2007}.

The adjacency is fixed throughout both training and inference. G-NAC
therefore does not rewire the graph. Instead, it learns recurrent
cellular states and time-varying message conductances on the fixed
neighborhood substrate.

Each undirected edge is stored in both directions for message
aggregation. Two distinct quantities are associated with each edge.
First, a fixed Gaussian edge weight $w_{ij}$ is computed from the
Euclidean input-space distance and is used by the graph-smoothness
regularizer:
\begin{equation}
    d_{ij}
    =
    \|x_i-x_j\|_2,
\end{equation}
\begin{equation}
    \sigma
    =
    \operatorname{median}
    \left(
        \{d_{ij}:(i,j)\in E_{\mathrm{dir}}\}
    \right),
\end{equation}
and
\begin{equation}
    w_{ij}
    =
    \exp
    \left(
        -\frac{d_{ij}^{2}}{2\sigma^{2}}
    \right).
\end{equation}

Second, a three-dimensional geometry vector $g_{ij}$ is supplied to
the learned cellular transition rule:
\begin{equation}
    g_{ij}
    =
    \begin{bmatrix}
        \widetilde{d}_{ij} &
        J_{ij} &
        M_{ij}
    \end{bmatrix},
\end{equation}
where $\widetilde{d}_{ij}$ is a locally normalized Euclidean distance,
$J_{ij}$ is the Jaccard overlap between the original directed kNN
neighbor sets, and $M_{ij}$ indicates whether the original neighbor
relation is mutual.

For each node, let
\begin{equation}
    s_i
    =
    \operatorname{median}
    \left(
        \{d_{i\ell}:\ell\in\mathcal{N}_{k}(i)\}
    \right).
\end{equation}
The locally normalized distance is
\begin{equation}
    \widetilde{d}_{ij}
    =
    \frac{d_{ij}}
    {\frac{1}{2}(s_i+s_j)+10^{-8}}.
\end{equation}
The neighborhood-overlap feature is
\begin{equation}
    J_{ij}
    =
    \frac{
        |\mathcal{N}_{k}(i)\cap\mathcal{N}_{k}(j)|
    }{
        |\mathcal{N}_{k}(i)\cup\mathcal{N}_{k}(j)|
    },
\end{equation}
and the mutual-neighbor indicator is
\begin{equation}
    M_{ij}
    =
    \mathbb{I}
    \left[
        j\in\mathcal{N}_{k}(i)
        \land
        i\in\mathcal{N}_{k}(j)
    \right].
\end{equation}

The fixed graph weight $w_{ij}$ should be distinguished from the
learned conductance introduced below. The former represents
input-graph similarity and is used for regularization, whereas the
latter is a time-dependent quantity learned by the cellular transition
rule and controls recurrent message exchange.

\subsection{Graph Neural Cellular Dynamics}
\label{sec:dynamics}

G-NAC builds upon the Graph Neural Cellular Automata (GNCA)
formulation of \citet{Grattarola2021}, in which a graph neural
network parameterizes a shared cellular transition rule over an
arbitrary graph. As in a GNCA, the same learned transition rule
is reused across graph cells and recurrent steps, allowing global
organization to emerge through repeated local interactions.

Each cell maintains the immutable observed feature vector $x_i$, an
internal recurrent hidden state
\begin{equation}
    h_i^{(t)}\in\mathbb{R}^{d_h},
\end{equation}
and an explicit domain state
\begin{equation}
    z_i^{(t)}\in\mathbb{R}^{d_z}.
\end{equation}
The complete cellular state can therefore be represented as
\begin{equation}
    s_i^{(t)}
    =
    \left(
        x_i,
        h_i^{(t)},
        z_i^{(t)}
    \right).
\end{equation}

The hidden state serves as internal computational memory for the
transition rule, whereas the domain state is the latent
representation whose relational organization is subsequently used for
clustering.

\subsubsection{State Initialization}
\label{sec:init}

At the beginning of each training rollout, a fresh Gaussian seed
\begin{equation}
    \xi_i\sim\mathcal{N}(0,I),
    \qquad
    \xi_i\in\mathbb{R}^{d_\xi},
\end{equation}
is sampled independently for every cell. This weak stochastic
initialization provides symmetry breaking between otherwise similar
cellular configurations.

The initial hidden state is
\begin{equation}
    h_i^{(0)}
    =
    \tanh
    \left(
        f_x(x_i)
        +
        s_{\xi}f_{\xi}(\xi_i)
    \right),
\end{equation}
where $f_x$ and $f_{\xi}$ are learned encoders and $s_{\xi}$ controls
the contribution of the stochastic seed.

The initial domain state is generated from the concatenated
observation and scaled seed:
\begin{equation}
    z_i^{(0)}
    =
    f_{z,0}
    \left(
        [x_i,s_{\xi}\xi_i]
    \right).
\end{equation}
The domain initializer has no final squashing activation, allowing the
anti-collapse objective to establish the scale of the domain
coordinates.

\subsubsection{Learned Edge Conductance and Message Construction}
\label{sec:conductance}

For every stored directed edge $i\rightarrow j$, G-NAC compares the
endpoint observations, hidden states, domain states, and fixed edge
geometry. The edge representation is
\begin{equation}
    e_{ij}^{(t)}
    =
    \left[
        |x_i-x_j|,
        |h_i^{(t)}-h_j^{(t)}|,
        |z_i^{(t)}-z_j^{(t)}|,
        g_{ij}
    \right].
\end{equation}

A learned edge network maps this representation to a scalar
conductance:
\begin{equation}
    \alpha_{ij}^{(t)}
    =
    \operatorname{sigmoid}
    \left(
        f_e(e_{ij}^{(t)})
    \right).
\end{equation}
The conductance is learned entirely through the unsupervised
objective; it is not trained as a supervised boundary classifier.

Conductance-weighted signed disagreement messages are then formed for
the hidden and domain states:
\begin{equation}
    m_{ij,h}^{(t)}
    =
    \alpha_{ij}^{(t)}
    \left(
        h_i^{(t)}-h_j^{(t)}
    \right),
\end{equation}
\begin{equation}
    m_{ij,z}^{(t)}
    =
    \alpha_{ij}^{(t)}
    \left(
        z_i^{(t)}-z_j^{(t)}
    \right).
\end{equation}

These messages are mean-aggregated at the destination node:
\begin{equation}
    \overline{m}_{j,h}^{(t)}
    =
    \operatorname{mean}_{i:(i,j)\in E_{\mathrm{dir}}}
    m_{ij,h}^{(t)},
\end{equation}
\begin{equation}
    \overline{m}_{j,z}^{(t)}
    =
    \operatorname{mean}_{i:(i,j)\in E_{\mathrm{dir}}}
    m_{ij,z}^{(t)}.
\end{equation}

\subsubsection{Recurrent Hidden and Domain Updates}
\label{sec:updates}

The hidden-state update receives the current hidden state, aggregated
hidden disagreement, current domain state, and immutable observation:
\begin{equation}
    \Delta h_j^{(t)}
    =
    f_h
    \left(
        [
        h_j^{(t)},
        \overline{m}_{j,h}^{(t)},
        z_j^{(t)},
        x_j
        ]
    \right).
\end{equation}
The resulting residual update is
\begin{equation}
    h_j^{(t+1)}
    =
    \tanh
    \left(
        h_j^{(t)}
        +
        \eta\Delta h_j^{(t)}
    \right),
\end{equation}
where $\eta$ is the recurrent update scale.

The domain update uses the newly updated hidden state, current domain
state, aggregated domain disagreement, and observation:
\begin{equation}
    \Delta z_j^{(t)}
    =
    f_z
    \left(
        [
        h_j^{(t+1)},
        z_j^{(t)},
        \overline{m}_{j,z}^{(t)},
        x_j
        ]
    \right),
\end{equation}
followed by
\begin{equation}
    z_j^{(t+1)}
    =
    z_j^{(t)}
    +
    \eta\Delta z_j^{(t)}.
\end{equation}

Both update networks terminate in hyperbolic-tangent activations,
thereby bounding each proposed update. The hidden state is additionally
passed through an outer $\tanh$ following the residual update. The
accumulated domain state is not externally squashed, allowing its
coordinate scale to evolve under the anti-collapse objective.

The same functions $f_e$, $f_h$, and $f_z$, and therefore the same
learned parameters, are applied to every graph cell and at every
recurrent step. G-NAC consequently learns a shared local rule of
cellular evolution rather than node-specific parameters or directly
supervised cluster assignments.

\subsection{Unsupervised Domain-Formation Objective}
\label{sec:objective}

G-NAC is trained without ground-truth cluster assignments. Its
objective combines local graph coherence, anti-collapse
regularization, weak non-local dispersion, and late-rollout temporal
stabilization:
\begin{equation}
\label{eq:total_loss}
    \mathcal{L}
    =
    \lambda_s\mathcal{L}_{\mathrm{smooth}}
    +
    \lambda_v\mathcal{L}_{\mathrm{var}}
    +
    \lambda_c\mathcal{L}_{\mathrm{cov}}
    +
    \lambda_u\mathcal{L}_{\mathrm{unif}}
    +
    \lambda_t\mathcal{L}_{\mathrm{temp}}.
\end{equation}

\subsubsection{Weighted Graph Smoothness}

Graph-based learning commonly assumes that strongly connected
observations should receive similar representations or function
values. Weighted squared differences over graph edges are a standard
graph-Laplacian smoothness construction
\citep{Belkin2006,Zhu2003}. G-NAC applies this principle to the final
domain state of the training rollout:
\begin{equation}
    \mathcal{L}_{\mathrm{smooth}}
    =
    \frac{
        \displaystyle
        \sum_{(i,j)\in E_{\mathrm{dir}}}
        w_{ij}
        \|z_i-z_j\|_2^2
    }{
        \displaystyle
        \sum_{(i,j)\in E_{\mathrm{dir}}}
        w_{ij}
        +
        10^{-8}
    }.
\end{equation}

Because each undirected edge is stored in both directions with the
same weight, the directed storage does not alter the normalized value
relative to summing once per undirected edge.

In isolation, graph smoothness admits the trivial globally collapsed
solution
\begin{equation}
    z_1=z_2=\cdots=z_N.
\end{equation}
The remaining objective terms therefore provide complementary
anti-collapse and differentiation pressures.

\subsubsection{Variance Regularization}

To penalize dimensional collapse, G-NAC uses a variance regularizer
inspired by Variance-Invariance-Covariance Regularization (VICReg)
\citep{Bardes2022}. For domain coordinate $q$, the implementation
computes the population variance across the $N$ cells:
\begin{equation}
\sigma_q
=
\sqrt{
    \operatorname{Var}_{\mathrm{pop}}(z_{\cdot q})
    +
    10^{-4}
}.
\end{equation}

The variance loss is
\begin{equation}
    \mathcal{L}_{\mathrm{var}}
    =
    \frac{1}{d_z}
    \sum_{q=1}^{d_z}
    \max(0,\gamma-\sigma_q),
\end{equation}
where $\gamma$ denotes the target minimum standard deviation.

The term penalizes coordinate standard deviations below $\gamma$ but
does not impose a hard variance floor or guarantee avoidance of collapsed
stationary states. It also does not reward coordinates for increasing
their standard deviation beyond $\gamma$.

\subsubsection{Covariance Regularization}

Variance alone does not ensure that different domain coordinates encode
distinct information. G-NAC therefore incorporates a VICReg-inspired
covariance penalty \citep{Bardes2022}. Let
\begin{equation}
    Z_c
    =
    Z-\overline{Z},
\end{equation}
where $\overline{Z}$ contains the coordinate-wise means. The sample
covariance matrix is
\begin{equation}
    \Sigma_z
    =
    \frac{Z_c^{\top}Z_c}{N-1}.
\end{equation}

G-NAC penalizes the mean squared off-diagonal covariance:
\begin{equation}
    \mathcal{L}_{\mathrm{cov}}
    =
    \frac{
        \displaystyle
        \sum_{p\neq q}
        (\Sigma_z)_{pq}^{2}
    }{
        d_z(d_z-1)
    }.
\end{equation}

This differs slightly in scaling from the original VICReg covariance
penalty: G-NAC averages over the off-diagonal entries rather than
normalizing only by the representation dimension.

\subsubsection{Non-Neighbor Dispersion}

G-NAC additionally applies a weak pairwise dispersion term inspired by
the Gaussian-potential uniformity objective of
\citet{WangIsola2020}. Their formulation encourages uniformity among
independently sampled, unit-normalized representations on the
hypersphere. G-NAC instead adapts the Gaussian-potential principle to
the relational structure of the graph.

Let
\begin{equation}
    \mathcal{P}_{\mathrm{NN}}
    \subset
    \left\{
        (i,j):
        i\neq j,\;
        j\notin\mathcal{N}_{G}(i)
    \right\}
\end{equation}
denote a randomly sampled set of explicit graph non-neighbor pairs.
The G-NAC loss is
\begin{equation}
    \mathcal{L}_{\mathrm{unif}}
    =
    \frac{1}{|\mathcal{P}_{\mathrm{NN}}|}
    \sum_{(i,j)\in\mathcal{P}_{\mathrm{NN}}}
    \exp
    \left(
        -\tau
        \|z_i-z_j\|_2^2
    \right).
\end{equation}

This term is not identical to the uniformity loss of Wang and Isola.
G-NAC does not constrain $z$ to the unit hypersphere,
omits the outer logarithm, and samples graph non-neighbors rather than
independent global pairs. Consequently, the term is more precisely
interpreted as a weak non-neighbor dispersion pressure than as
hyperspherical uniformity.

Importantly, non-neighboring cells are not assumed to belong to
different ground-truth clusters. Two observations belonging to the
same cluster or manifold may be sufficiently distant that they are
not directly connected in the kNN graph. The term is therefore
assigned a comparatively small coefficient and serves only as a weak
non-local anti-collapse pressure.

\subsubsection{Temporal Domain Stability}

Because G-NAC defines clustering through recurrent dynamics, the
training objective additionally encourages the final portion of each
rollout to change gradually. For a sequence of domain states near the
end of the rollout,
\begin{equation}
    \mathcal{L}_{\mathrm{temp}}
    =
    \frac{1}{|\mathcal{T}|}
    \sum_{t\in\mathcal{T}}
    \operatorname{MSE}
    \left(
        z^{(t+1)},
        \operatorname{sg}
        \left[
            z^{(t)}
        \right]
    \right),
\end{equation}
where $\operatorname{sg}[\cdot]$ denotes stop-gradient.

The preceding state is therefore treated as a detached target for the
current state. This discourages abrupt late-rollout changes without
forcing earlier cellular states to remain static.

With the benchmark configuration of 16 recurrent training steps and
four retained tail states, the temporal loss contains the three
successive transitions between those final four states. This training
regularizer is distinct from the Spearman rank-stability criterion
used for adaptive inference.

\subsubsection{Objective Interpretation}

The complete objective can be interpreted as a competition between
local attraction and global differentiation:
\begin{equation}
\begin{split}
    \mathcal{L}_{\mathrm{smooth}}
        &\quad\rightarrow\quad \text{local coherence},\\
    \mathcal{L}_{\mathrm{var}}
        &\quad\rightarrow\quad \text{anti-collapse},\\
    \mathcal{L}_{\mathrm{cov}}
        &\quad\rightarrow\quad \text{coordinate decorrelation},\\
    \mathcal{L}_{\mathrm{unif}}
        &\quad\rightarrow\quad \text{weak non-local dispersion},\\
    \mathcal{L}_{\mathrm{temp}}
        &\quad\rightarrow\quad \text{late-rollout stabilization}.
\end{split}
\end{equation}

Graph smoothness encourages neighboring cells to organize into similar
domain states, but alone favors global consensus. Variance and
non-neighbor dispersion oppose this collapse, while covariance
regularization discourages redundant latent coordinates. The desired
configuration is therefore neither global collapse nor unrestricted
separation, but a set of \emph{locally coherent and globally
differentiated domains}.

\subsection{Training and Unsupervised Checkpoint Selection}
\label{sec:training}

At every training epoch, G-NAC samples a fresh Gaussian cellular seed,
initializes the hidden and domain states, and unrolls the same shared
transition rule for a fixed number of recurrent steps. Graph
non-neighbor pairs are sampled anew for the weak dispersion term. The
five loss components in Eq.~\eqref{eq:total_loss} are then evaluated
and differentiated through the recurrent rollout.

Optimization uses AdamW with gradient-norm clipping. Model selection
remains entirely unsupervised. Following a burn-in period, the
parameter state having the lowest total unsupervised training loss is
retained. Ground-truth labels and external clustering metrics such as
ARI, AMI, and NCA are not used for training or checkpoint selection.

For the publication benchmark, the frozen configuration is summarized
in Table~\ref{tab:gnac_config}.

\begin{table}[t]
    \centering
    \caption{Frozen G-NAC configuration used in the publication benchmark.}
    \label{tab:gnac_config}
    \begin{tabular}{lll}
        \toprule
        Parameter & Value & Role \\
        \midrule
        $k$ & 20 & kNN graph degree \\
        Training rollout & 16 & Recurrent updates per epoch \\
        Training epochs & 100 & Benchmark training budget \\
        Learning rate & $2\times10^{-3}$ & AdamW learning rate \\
        Weight decay & $1\times10^{-5}$ & AdamW weight decay \\
        Gradient clip & 5.0 & Maximum gradient norm \\
        $\lambda_s$ & 1.0 & Graph smoothness \\
        $\lambda_v$ & 4.0 & Variance anti-collapse \\
        $\lambda_c$ & 0.10 & Covariance regularization \\
        $\lambda_u$ & 0.10 & Non-neighbor dispersion \\
        $\lambda_t$ & 0.05 & Temporal stabilization \\
        $\gamma$ & 1.0 & Target standard deviation \\
        Non-neighbor pairs & 3000 & Sampled per epoch \\
        $\tau$ & 2.0 & Dispersion temperature \\
        Temporal tail & 4 & Retained final states \\
        $d_h$ & 48 & Hidden-state dimension \\
        $d_z$ & 8 & Domain-state dimension \\
        $d_\xi$ & 4 & Stochastic-seed dimension \\
        $\eta$ & 0.15 & Recurrent update scale \\
        $s_\xi$ & 0.10 & Seed scale \\
        Burn-in & 40 epochs & Checkpoint eligibility \\
        \bottomrule
    \end{tabular}
\end{table}

All benchmark experiments use a fixed training budget of 100 epochs
unless otherwise stated.

\subsection{Relational Stability and Adaptive Inference}
\label{sec:stopping}

After training, the best unsupervised parameter checkpoint is restored
and the cellular rule is evaluated from a deterministic inference
seed. 

Rather than requiring the absolute domain coordinates to approach a
fixed point, G-NAC measures stabilization of the relative organization
of graph edges. At recurrent checkpoint $T$, Euclidean domain-space
distances are computed over the unique undirected graph-edge set
$E_u$:
\begin{equation}
    D^{(T)}
    =
    \left[
        \|z_i^{(T)}-z_j^{(T)}\|_2
    \right]_{\{i,j\}\in E_u}.
\end{equation}

Relational stability is measured by the Spearman rank correlation
between the edge-distance vectors at successive checkpoints:
\begin{equation}
    \rho_T
    =
    \rho_S
    \left(
        D^{(T_{\mathrm{prev}})},
        D^{(T)}
    \right).
\end{equation}
The implementation passes the distance vectors directly to Spearman's
statistic; the statistic itself evaluates the correlation between
their ranks.

The frozen checkpoint schedule is
\begin{equation}
    \mathcal{C}
    =
    \{2,4,8,16,32,64,128\}.
\end{equation}
The first eligible stopping decision occurs at $T=4$, comparing the
states at $T=2$ and $T=4$. The selected recurrent depth is
\begin{equation}
    T^{*}
    =
    \min
    \left\{
        T\in\mathcal{C}:
        T\geq4,\;
        \rho_T\geq0.95
    \right\}.
\end{equation}
If no checkpoint satisfies the criterion, $T^{*}=128$.

This criterion is used as an empirical \emph{rank-stability stopping
rule}: inference may terminate when successive checkpoint edge-distance
rankings are highly correlated even if the latent coordinates continue
to change in absolute scale or position. A single threshold crossing does
not guarantee future convergence, stability of every individual edge, or
partition invariance at later rollout depths.

The G-NAC implementation may either compute the complete rollout
to $T=128$ and subsequently select $T^{*}$, or exit when the same
stopping condition is satisfied. This implementation-level optimization 
changes the computational cost but not the mathematical stopping rule.

\subsection{Global Edge-Rank Affinity and Spectral Readout}
\label{sec:readout}

G-NAC is trained to organize relationships between graph-connected
cells rather than to impose a prescribed metric scale on the latent
space. The final readout therefore uses the relative ordering of
domain-space edge distances rather than their absolute magnitudes.
This is related to previous rank-based clustering and spectral
affinity methods \citep{Zhu2011RankOrder,YeSakurai2016}, but G-NAC
ranks learned domain-space distances globally over the edges of the
fixed graph.

Let $E_u$ denote the set of $M=|E_u|$ unique undirected graph edges.
At the selected inference step $T^*$, define
\begin{equation}
    d_e
    =
    \|z_i^{(T^{*})}-z_j^{(T^{*})}\|_2,
    \qquad e=\{i,j\}\in E_u.
\end{equation}
Each edge is assigned its global ordinal rank
\begin{equation}
    r_e
    =
    \operatorname{rank}
    \left(
        d_e;
        \{d_{e'}\}_{e'\in E_u}
    \right),
    \qquad
    r_e\in\{0,\ldots,M-1\},
\end{equation}
with the smallest distance assigned rank zero. For $M>1$,
\begin{equation}
    q_e=\frac{r_e}{M-1},
\end{equation}
while $q_e=0$ for $M=1$. The corresponding affinity is
\begin{equation}
\label{eq:rank_affinity}
    a_e
    =
    \max\left(1-q_e,10^{-6}\right).
\end{equation}

The implementation obtains ordinal ranks using a stable ascending
sort. Exact distance ties are therefore resolved by deterministic edge
ordering rather than assigned a common average rank. Average-rank tie
handling would provide a permutation-compatible alternative. Subject
to fixed edge support and the same tie convention, the affinity is
invariant to strictly increasing transformations of the edge
distances; changes in latent scale that preserve their ordering
therefore leave the affinity unchanged.

The sparse symmetric affinity matrix $A^{(Z)}$ retains the support of
the original neighborhood graph and uses unit diagonal affinities,
$A_{ii}^{(Z)}=1$. Before spectral partitioning, G-NAC checks the
connected components of its off-diagonal support. If the graph contains
exactly $K$ components, their labels are returned directly. Otherwise,
normalized spectral clustering is applied to $A^{(Z)}$ as a
precomputed similarity matrix \citep{vonLuxburg2007}, followed by
$K$-means with 20 initializations. Thus,
\begin{equation}
    \mathcal{R}_{K}(A)
    =
    \begin{cases}
        \operatorname{Components}(A), & c(A)=K,\\
        \operatorname{Spectral}(A,K), & c(A)\neq K.
    \end{cases}
\end{equation}

The requested cluster count $K$ therefore enters only through the
final readout $\mathcal{R}_{K}$, not during training, checkpoint
selection, or recurrent inference. Both adaptive stopping and affinity
construction depend on the ordering of graph-edge distances: the former
detects high rank agreement between successive checkpoints, while the
latter converts the selected ordering into the affinity used for
partitioning.

\subsection{Algorithm Summary}
\label{sec:algorithm}

\begin{figure}[htbp]
    \centering
    \includegraphics[width=1\linewidth]{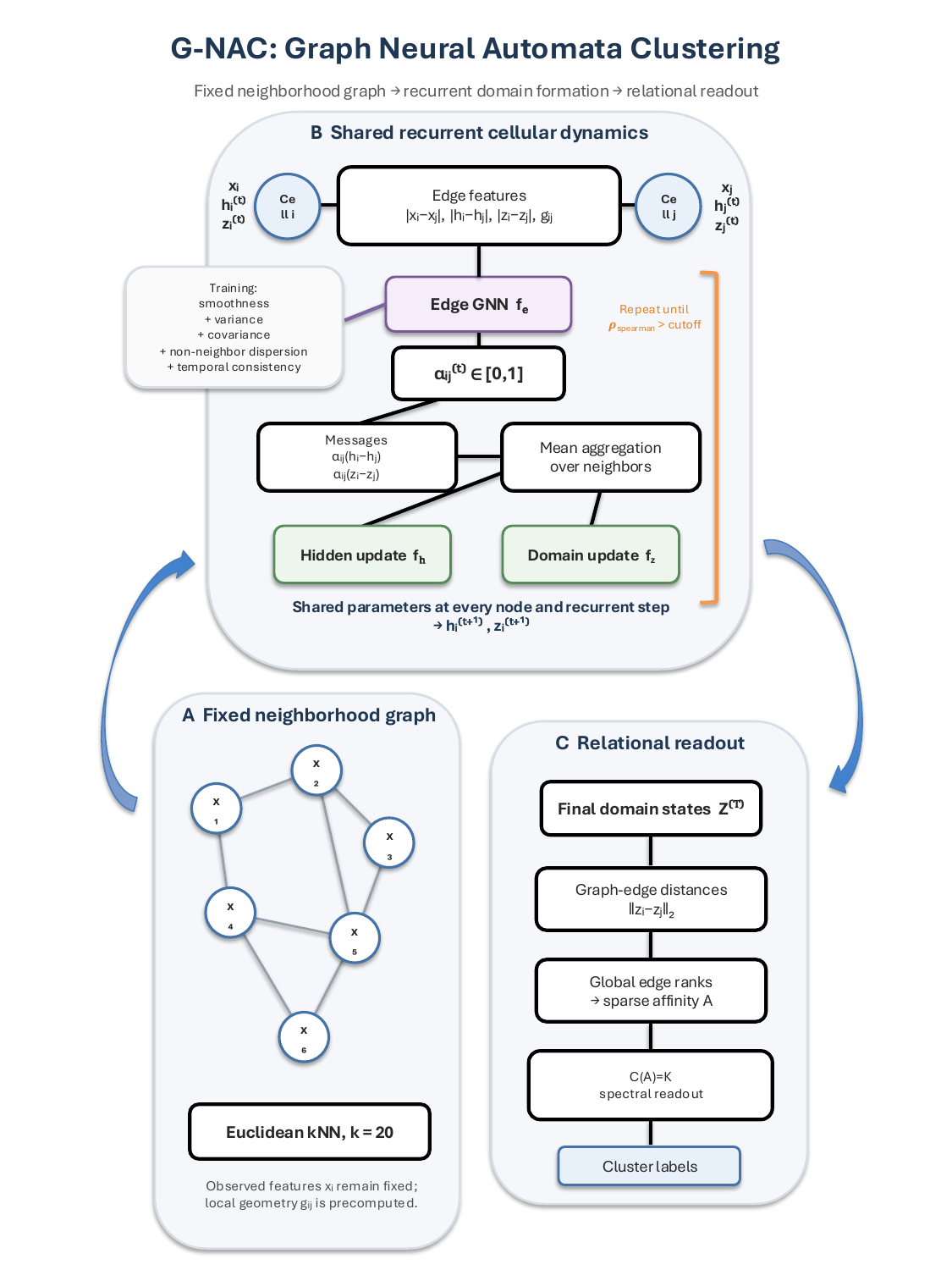}
    \caption{Detailed G-NAC model architecture and recurrent information flow.}
    \label{fig:gnac-architecture}
\end{figure}

Algorithm~\ref{alg:gnac} summarizes the complete G-NAC procedure. Figure~\ref{fig:gnac-architecture} provides the corresponding detailed model architecture, complementing the conceptual workflow in Figure~\ref{fig:conceptual_comparison}.

\begin{algorithm}[H]
\caption{Graph Neural Automata Clustering (G-NAC)}
\label{alg:gnac}
\begin{algorithmic}[1]
\Require Feature matrix $X$, requested cluster count $K$, graph degree $k$
\Ensure Cluster assignments $C$

\State Construct Euclidean $k$NN graph $G=(V,E)$
\State Symmetrize the directed neighbor relation by union
\State Compute fixed smoothness weights $w_{ij}$ and geometry $g_{ij}$
\State Initialize shared GNCA parameters $\theta$

\For{$e=1,\ldots,E_{\mathrm{train}}$}
    \State Sample fresh cellular seeds $\xi_i\sim\mathcal{N}(0,I)$
    \State Initialize $H^{(0)}$ and $Z^{(0)}$ from $X$ and $\xi$

    \For{$t=0,\ldots,T_{\mathrm{train}}-1$}
        \ForAll{directed edges $i\rightarrow j$}
            \State $e_{ij}\gets
            [|x_i-x_j|,|h_i-h_j|,|z_i-z_j|,g_{ij}]$
            \State $\alpha_{ij}\gets
            \operatorname{sigmoid}(f_e(e_{ij}))$
            \State $m_{ij,h}\gets\alpha_{ij}(h_i-h_j)$
            \State $m_{ij,z}\gets\alpha_{ij}(z_i-z_j)$
        \EndFor

        \State Mean-aggregate incoming messages at each destination
        \State Update $H^{(t+1)}$ using the shared hidden rule
        \State Update $Z^{(t+1)}$ using the shared domain rule
    \EndFor

    \State Sample graph non-neighbor pairs $\mathcal{P}_{\mathrm{NN}}$
    \State Compute $\mathcal{L}_{\mathrm{smooth}}$,
           $\mathcal{L}_{\mathrm{var}}$,
           $\mathcal{L}_{\mathrm{cov}}$,
           $\mathcal{L}_{\mathrm{unif}}$, and
           $\mathcal{L}_{\mathrm{temp}}$
    \State Zero optimizer gradients
    \State Compute total objective using Eq.~\eqref{eq:total_loss}
    \State Backpropagate through the recurrent rollout
    \State Clip the global gradient norm to 5
    \State Apply one AdamW optimizer step

    \If{burn-in is complete and total loss is the lowest observed}
        \State Store parameter checkpoint $\theta^{*}$
    \EndIf
\EndFor

\State Restore $\theta^{*}$
\State Initialize cellular states from deterministic inference seed

\For{$T\in\{2,4,8,16,32,64,128\}$}
    \State Evolve the shared GNCA rule to checkpoint $T$
    \State Compute unique-edge latent-distance vector $D^{(T)}$

    \If{$T\geq4$}
        \State $\rho_T\gets
        \rho_S(D^{(T_{\mathrm{prev}})},D^{(T)})$

        \If{$\rho_T\geq0.95$}
            \State $T^{*}\gets T$
            \State \textbf{break}
        \EndIf
    \EndIf
    \State $T_{\mathrm{prev}}\gets T$
\EndFor

\If{no checkpoint satisfied the stopping criterion}
    \State $T^{*}\gets128$
\EndIf

\State Rank unique graph-edge distances at $Z^{(T^{*})}$
\State Convert ranks to affinities using Eq.~\eqref{eq:rank_affinity}
\State Form symmetric sparse affinity matrix $A^{(Z)}$
\State Set $A_{ii}^{(Z)}\gets1$
\State $(c,\ell)\gets\operatorname{ConnectedComponents}(A^{(Z)})$
\If{$c=K$}
    \State $C\gets\ell$ \Comment{component-resolved readout}
\Else
    \State $C\gets\operatorname{SpectralClustering}(A^{(Z)},K)$
\EndIf
\State \Return $C$

\end{algorithmic}
\end{algorithm}


\section{Datasets and Experimental Tasks}
\label{sec:data_tasks}

\subsection{Benchmark Collections}
\label{sec:benchmark_design}

The primary benchmark uses the versioned Clustering Benchmarks suite
\citep{Gagolewski2022Benchmark,Gagolewski2022BenchmarkSuite}. It
contains 73 dataset/$K$ tasks drawn from 57 datasets in four batteries:
the Fundamental Clustering and Projection Suite (FCPS)
\citep{UltschLoetsch2020FCPS}, the SIPU benchmark collection
\citep{FrantiSieranoja2018}, the Graves collection
\citep{GravesPedrycz2010}, and the WUT collection. WUT is distributed
as part of the same versioned Clustering Benchmarks suite rather than
cited here as a separate standalone source. The benchmark observations
are numerical feature vectors rather than application-specific images,
signals, or class decisions. Across the retained tasks, datasets contain
120--10,000 observations with two or three observed features, and the
reference partitions contain between 2 and 50 clusters. Table~\ref{tab:benchmark_composition}
summarizes the composition by battery.

\begin{table}[H]
    \centering
    \caption{Composition of the primary ClustBench evaluation. Ranges are
    computed over the retained dataset/$K$ tasks used in the study.}
    \label{tab:benchmark_composition}
    \begin{tabular}{lrrrrr}
        \toprule
        Battery & Datasets & Tasks & $N$ range & $d$ range & $K$ range \\
        \midrule
        FCPS   & 9  & 10 & 212--4096   & 2--3 & 2--7 \\
        SIPU   & 16 & 21 & 240--7500   & 2    & 2--50 \\
        Graves & 10 & 17 & 200--1050   & 2    & 2--5 \\
        WUT    & 22 & 25 & 120--10000  & 2--3 & 2--10 \\
        \midrule
        Total  & 57 & 73 & 120--10000  & 2--3 & 2--50 \\
        \bottomrule
    \end{tabular}
\end{table}

A benchmark \emph{task} consists of an unlabeled feature matrix $X$ and
a requested cluster count $K$. A method returns one cluster assignment
for every observation. The reference partition is withheld from fitting
and used only after clustering to compute external agreement metrics.
Thus, the benchmark is an unsupervised partition-recovery problem rather
than supervised classification, and there is no supervised training/test
split for the 73 benchmark tasks. When a dataset contains more than one
valid target value of $K$, each dataset/$K$ combination is treated as a
separate task.

The requested $K$ is obtained from the benchmark reference-partition
metadata and supplied to every method that requires a cluster count. The
reference assignments themselves are never provided to G-NAC during
training, checkpoint selection, or recurrent inference. All methods
receive the same feature representation for a given task.

\subsection{Benchmark Methods and Performance Measures}

Eleven methods are evaluated: G-NAC; $K$-means
\citep{MacQueen1967KMeans}; Gaussian mixture modeling (GMM), fit by the
expectation--maximization framework \citep{Dempster1977EM}; Ward and
average-linkage hierarchical clustering \citep{Ward1963,Jain2010DataClustering};
BIRCH \citep{Zhang1996BIRCH}; spectral clustering
\citep{vonLuxburg2007}; Genie \citep{Gagolewski2016Genie}; DEC
\citep{Xie2016DEC}; IDEC \citep{Guo2017IDEC}; and SOM
\citep{Kohonen1982SOM}. These baselines span centroid-based,
probabilistic, hierarchical, graph-spectral, self-organizing, and deep
embedded clustering approaches.

Performance is quantified using Adjusted Rand Index (ARI)
\citep{HubertArabie1985}, Adjusted Mutual Information (AMI)
\citep{Vinh2010AMI}, and Normalized Clustering Accuracy (NCA) as
implemented by \texttt{genieclust} and used in the benchmark suite
\citep{Gagolewski2022Benchmark}. These metrics compare the recovered
partition with the withheld reference partition after clustering; they
do not participate in G-NAC optimization or checkpoint selection.

\subsection{Controlled Scaling and Transfer Tasks}

The robustness experiment reuses the same 57 datasets and 73
benchmark tasks while perturbing either the observed features or graph
relations. The scalability and transfer experiments are separate
controlled synthetic studies. Scalability uses a balanced $K=8$,
$d=16$ isotropic Gaussian-mixture problem with graph size varied from
1,000 to 100,000 observations. Sparse-to-full transfer uses two
$d=16$, $K=8$ synthetic families---a Gaussian mixture and noisy
blobs---with independently sampled source and 100,000-node target
graphs. Unlike the primary benchmark, this transfer experiment therefore
has an explicit source/target distinction: model parameters are learned
on the source graph, frozen, and then evaluated on an independently
sampled target graph from the same generating process.


\section{Experimental Evaluation}
\label{sec:experiments}

We evaluate G-NAC through four experiments. First, a multi-battery
benchmark compares clustering performance and stochastic stability
against ten classical, spectral, self-organizing, and deep clustering
baselines. Second, measurement-noise and graph-corruption experiments
examine dependence on the observed features and local interaction
topology. Third, a controlled scaling experiment measures runtime and
memory requirements from $10^3$ to $10^5$ graph cells. Finally, a
sparse-to-full transfer experiment tests whether cellular rules learned
on smaller source graphs can be deployed without retraining on
independent 100,000-node target graphs.

\subsection{Experimental Protocol}
\label{sec:protocol}

Stochastic methods were evaluated using ten algorithm seeds,
\begin{equation}
    \mathcal{S}
    =
    \{7,17,27,37,47,57,67,77,87,97\},
\end{equation}
while deterministic methods were evaluated once per dataset/$K$ task.
G-NAC used the fixed configuration in Table~\ref{tab:gnac_config}
throughout: $k=20$, a 16-step training rollout, 100 epochs, and the
relational stopping rule of Section~\ref{sec:stopping}. No
benchmark-specific G-NAC tuning was performed.

ARI, AMI, and NCA are defined in Section~\ref{sec:data_tasks}, with
ARI used as the primary metric. For datasets with multiple valid same-$K$
reference partitions, the maximum agreement over compatible references
was computed separately for each metric. Runtime denotes total wall-clock
execution time and is reported as an empirical cost measure rather than
a hardware-independent complexity comparison.

\subsection{Statistical Analysis}
\label{sec:statistics}

Descriptive summaries use the 73 dataset/$K$ tasks, with stochastic
runs first averaged within each task. Because some underlying datasets
contribute multiple values of $K$, inferential comparisons instead use
the dataset as the independent unit.

For baseline $b$, task-level paired differences are
\begin{equation}
    \Delta_t^{(b)}
    =
    \operatorname{ARI}_{t}^{\mathrm{G\text{-}NAC}}
    -
    \operatorname{ARI}_{t}^{(b)}.
\end{equation}
Multiple task effects from the same dataset were averaged before
inference. Two-sided paired Wilcoxon signed-rank tests were applied to
the resulting dataset-level scores, with Holm correction across the ten
G-NAC--baseline comparisons. Confidence intervals for mean ARI
differences were obtained by nonparametric bootstrap resampling of
datasets.

Thus, benchmark means, medians, and win/tie/loss counts are descriptive
task-level quantities, whereas confidence intervals and hypothesis
tests use the underlying dataset as the independent unit.

\subsection{Overall Benchmark Performance}
\label{sec:overall_results}

Table~\ref{tab:method_summary} summarizes performance across the
73 dataset/$K$ tasks. G-NAC achieved the highest mean and median ARI
among the eleven evaluated methods, with a mean ARI of
$0.7951$ and a median ARI of $0.9379$.

Genie produced nearly identical mean ARI, achieving $0.7941$, with a
median ARI of $0.9152$. The difference in mean ARI between G-NAC and
Genie was therefore only approximately
\begin{equation}
    0.795072 - 0.794053 = 0.001019.
\end{equation}

Spectral clustering was the next strongest method by mean ARI at
$0.7011$, followed by GMM at $0.6510$. The remaining methods obtained
mean ARI values between approximately $0.47$ and $0.56$.

\begin{table*}[t]
    \centering
    \caption{Aggregate clustering performance across the 73
    dataset/$K$ tasks in the benchmark experiment. Stochastic methods are
    first averaged across algorithm seeds within each task. Seed SD
    denotes the mean within-task standard deviation of ARI across
    stochastic seeds.}
    \label{tab:method_summary}
    \begin{tabular}{lrrrrrr}
        \toprule
        Method &
        Mean ARI &
        Median ARI &
        Mean AMI &
        Mean NCA &
        Seed SD &
        Time (s) \\
        \midrule
        G-NAC    & 0.7951 & 0.9379 & 0.8350 & 0.8463 & 0.0193 & 10.968 \\
        Genie    & 0.7941 & 0.9152 & 0.8503 & 0.8461 & 0.0000 & 0.005 \\
        Spectral & 0.7011 & 0.7668 & 0.7747 & 0.7840 & 0.0222 & 0.335 \\
        GMM      & 0.6510 & 0.8306 & 0.7024 & 0.7394 & 0.0300 & 0.055 \\
        $K$-means& 0.5580 & 0.5819 & 0.6256 & 0.6909 & 0.0022 & 0.070 \\
        IDEC     & 0.5572 & 0.5694 & 0.6294 & 0.6979 & 0.0637 & 3.189 \\
        Ward     & 0.5559 & 0.6015 & 0.6338 & 0.6898 & 0.0000 & 0.091 \\
        DEC      & 0.5388 & 0.5722 & 0.6190 & 0.6801 & 0.0575 & 2.716 \\
        SOM      & 0.5330 & 0.5771 & 0.6104 & 0.6668 & 0.0201 & 0.126 \\
        Average  & 0.5215 & 0.5197 & 0.6010 & 0.6549 & 0.0000 & 0.072 \\
        BIRCH    & 0.4691 & 0.4458 & 0.5709 & 0.6257 & 0.0000 & 0.014 \\
        \bottomrule
    \end{tabular}
\end{table*}

The three agreement metrics do not produce exactly the same ordering.
Although G-NAC achieved the highest mean ARI, Genie obtained the
highest mean AMI ($0.8503$ versus $0.8350$ for G-NAC). Their mean NCA
values were nearly identical ($0.8463$ for G-NAC and $0.8461$ for
Genie). These results indicate that the two methods exhibit comparable
overall benchmark performance while emphasizing slightly different
partition characteristics.

In contrast, G-NAC showed a considerably larger separation from
ordinary spectral clustering. Mean ARI increased from $0.7011$ for
spectral clustering to $0.7951$ for G-NAC, corresponding to an
absolute mean improvement of approximately $0.094$.

This comparison is particularly relevant because both methods operate
on graph structure. However, ordinary spectral clustering partitions
its similarity graph directly, whereas G-NAC evolves node states
through a learned recurrent cellular rule and then constructs a
rank-based sparse affinity for its general spectral readout. The
comparison therefore establishes a difference between the complete
G-NAC pipeline and the evaluated ordinary spectral baseline, but does
not by itself isolate the separate contributions of recurrent learning,
graph support, and rank-affinity construction.

\subsubsection{Paired Statistical Comparisons}
\label{sec:paired_results}

Aggregate rankings can obscure whether performance differences are
consistent across datasets. We therefore compared G-NAC against each
baseline using dataset-level paired ARI effects as described in
Section~\ref{sec:statistics}. Table~\ref{tab:paired_ari} reports the
mean dataset-level difference
\[
    \Delta\mathrm{ARI}
    =
    \mathrm{ARI}_{\mathrm{G\text{-}NAC}}
    -
    \mathrm{ARI}_{\mathrm{baseline}},
\]
together with dataset-bootstrap 95\% confidence intervals and
Holm-adjusted two-sided Wilcoxon signed-rank $p$-values. The
win/tie/loss counts remain descriptive counts over the 73
dataset/$K$ tasks.

\begin{table*}[t]
    \centering
    \caption{Paired ARI comparisons between G-NAC and the baseline
    methods. Positive $\Delta$ARI favors G-NAC. Stochastic runs are
    first averaged within each dataset/$K$ task; paired effects from
    multiple $K$ tasks are then averaged within each underlying dataset
    for inference. Confidence intervals are obtained by bootstrap
    resampling of datasets. Win/tie/loss counts are descriptive
    task-level counts. $p_{\mathrm{Holm}}$ denotes the Holm-adjusted
    two-sided Wilcoxon signed-rank $p$-value from the dataset-level
    analysis.}
    \label{tab:paired_ari}
    \begin{tabular}{lrrrr}
        \toprule
        Baseline &
        Mean $\Delta$ARI &
        95\% CI &
        Task W/T/L &
        $p_{\mathrm{Holm}}$ \\
        \midrule
        BIRCH
        & +0.3587
        & [0.2729, 0.4466]
        & 62/3/8
        & $1.31\times10^{-8}$ \\

        Average linkage
        & +0.2945
        & [0.2045, 0.3900]
        & 54/6/13
        & $7.06\times10^{-7}$ \\

        DEC
        & +0.2922
        & [0.2129, 0.3759]
        & 56/2/15
        & $1.59\times10^{-7}$ \\

        SOM
        & +0.2900
        & [0.2030, 0.3816]
        & 57/2/14
        & $1.59\times10^{-7}$ \\

        IDEC
        & +0.2679
        & [0.1902, 0.3497]
        & 60/0/13
        & $5.05\times10^{-8}$ \\

        Ward
        & +0.2649
        & [0.1771, 0.3589]
        & 53/6/14
        & $3.60\times10^{-7}$ \\

        $K$-means
        & +0.2627
        & [0.1733, 0.3586]
        & 50/7/16
        & $2.12\times10^{-6}$ \\

        GMM
        & +0.1651
        & [0.0785, 0.2597]
        & 41/9/23
        & 0.00131 \\

        Spectral
        & +0.0834
        & [0.0352, 0.1371]
        & 38/16/19
        & 0.00607 \\

        Genie
        & +0.00025
        & [-0.0695, 0.0634]
        & 34/20/19
        & 0.358 \\
        \bottomrule
    \end{tabular}
\end{table*}

After accounting for repeated cluster-count tasks from the same
underlying datasets, the inferential conclusions were unchanged for
the full benchmark. G-NAC differed significantly from nine of the ten
evaluated baselines after Holm correction. Relative to ordinary
spectral clustering, the dataset-level mean paired ARI difference was
$0.0834$ (95\% bootstrap CI $[0.0352,0.1371]$;
$p_{\mathrm{Holm}}=0.00607$).

Because the ordinary spectral baseline differs from G-NAC in more than
the recurrent state evolution, this significant whole-pipeline
difference should not be interpreted as a mechanism-isolating ablation.
It establishes that the complete G-NAC procedure differs from the
evaluated ordinary spectral baseline under the benchmark protocol;
matched graph/readout controls are required to attribute that
difference specifically to recurrence or learned conductance.

G-NAC also differed significantly from GMM in the full benchmark, with
a dataset-level mean paired improvement of $0.1651$ (95\% bootstrap CI
$[0.0785,0.2597]$; $p_{\mathrm{Holm}}=0.00131$).

In contrast, no statistically significant difference was observed
between G-NAC and Genie. Their dataset-level mean paired ARI difference
was $0.00025$, with a 95\% bootstrap confidence interval spanning zero
($[-0.0695,0.0634]$) and $p_{\mathrm{Holm}}=0.358$. At the descriptive
task level, G-NAC won 34 tasks, tied on 20, and lost 19. The near-zero
observed mean difference therefore does not establish statistical
equivalence; the interval remains compatible with non-negligible
differences in either direction.

\subsubsection{Performance Across Benchmark Batteries}
\label{sec:battery_results}

Table~\ref{tab:battery_summary} separates performance by benchmark
battery. G-NAC obtained the highest mean ARI on FCPS ($0.9530$), SIPU
($0.8074$), and Graves ($0.7953$), while Genie was highest on WUT
($0.7434$ versus $0.7214$ for G-NAC). G-NAC exceeded ordinary spectral
clustering in mean ARI across all four batteries, while the relative
ordering of G-NAC and Genie varied by collection.

\begin{table}[t]
    \centering
    \caption{Mean ARI by benchmark battery for the strongest methods.
    Values are averaged over dataset/$K$ task means within each
    battery.}
    \label{tab:battery_summary}
    \begin{tabular}{lrrrr}
        \toprule
        Method & FCPS & SIPU & Graves & WUT \\
        \midrule
        G-NAC
        & \textbf{0.9530}
        & \textbf{0.8074}
        & \textbf{0.7953}
        & 0.7214 \\

        Genie
        & 0.9249
        & 0.8004
        & 0.7837
        & \textbf{0.7434} \\

        Spectral
        & 0.8786
        & 0.6444
        & 0.7553
        & 0.6409 \\

        GMM
        & 0.7646
        & 0.7203
        & 0.5718
        & 0.6013 \\
        \bottomrule
    \end{tabular}
\end{table}

\subsubsection{Stochastic Stability}
\label{sec:seed_stability}

G-NAC exhibited a mean within-task ARI standard deviation of $0.0193$
across stochastic seeds, compared with $0.0222$ for spectral
clustering, $0.0300$ for GMM, $0.0575$ for DEC, $0.0637$ for IDEC, and
$0.0201$ for SOM. Thus, G-NAC showed relatively low, although nonzero,
sensitivity to stochastic initialization.

\subsubsection{Computational Cost}
\label{sec:runtime}

G-NAC incurred substantially greater computational cost than the
classical baselines, with mean task runtime $10.97$~s compared with
$0.005$~s for Genie, $0.335$~s for spectral clustering, $0.070$~s for
$K$-means, and $0.055$~s for GMM. It was also slower than DEC
($2.72$~s) and IDEC ($3.19$~s). Large-scale runtime and memory behavior
are examined separately in Section~\ref{sec:scalability_experiment}.

\subsection{Robustness to Measurement and Graph Corruption}
\label{sec:robustness_experiment}

The robustness experiment reused the 57 datasets and 73 dataset/$K$
tasks under additive Gaussian feature noise, random deletion of
directed $k$NN slots, and random rewiring of directed $k$NN slots at
\begin{equation}
    \mathcal{R}
    =
    \{0,0.05,0.10,0.20,0.30,0.50\}.
\end{equation}
Each condition used the same ten matched corruption and algorithm
seeds as the main benchmark.

Gaussian noise was scaled so that its expected RMS vector magnitude
was $r\,r_X$, where $r_X$ is the RMS Euclidean radius of the clean
sample. All methods received the same perturbed observations, and
graph-based methods rebuilt their graphs. For graph corruption, $X$
remained fixed while a fraction $r$ of each node's $k=20$ directed
neighbor slots was either deleted or replaced by sampled non-neighbors
before union symmetrization. Local distance scales and the clean RBF
scale remained fixed; overlap and mutual-neighbor attributes were
recomputed, no connectivity repair was applied, and G-NAC was
retrained for every condition.

Graph-corruption experiments additionally included a shared-support
spectral control using the same corrupted sparse support as G-NAC.
Replicates were averaged within dataset/$K$ tasks and repeated $K$
tasks within datasets were then averaged for dataset-level inference.
Bootstrap intervals and two-sided Wilcoxon tests were Holm-corrected
within each corruption family and metric. Table~\ref{tab:robustness_summary}
summarizes mean task-level ARI across the corruption conditions.

\begin{table*}[t]
    \centering
    \caption{Mean task-level ARI under measurement and graph corruption.
    Values first average the ten matched replicates within each of the 73
    dataset/$K$ tasks. The shared spectral control uses the same corrupted
    graph support as G-NAC.}
    \label{tab:robustness_summary}
    \small
    \resizebox{\textwidth}{!}{%
    \begin{tabular}{llrrrrrr}
        \toprule
        Corruption & Method & 0 & 0.05 & 0.10 & 0.20 & 0.30 & 0.50 \\
        \midrule
        Gaussian noise
            & G-NAC
            & 0.7951 & 0.7571 & 0.7090 & 0.5899 & 0.4786 & 0.3119 \\
        & Genie
            & 0.7941 & 0.7581 & 0.6781 & 0.5425 & 0.4289 & 0.2543 \\
        & Spectral
            & 0.7011 & 0.6790 & 0.6599 & 0.5793 & 0.4476 & 0.3034 \\
        & GMM
            & 0.6510 & 0.6392 & 0.6227 & 0.5405 & 0.4615 & 0.3353 \\
        \midrule
        Edge deletion
            & G-NAC
            & 0.7951 & 0.7944 & 0.7957 & 0.7966 & 0.7977 & 0.7990 \\
        & Shared spectral
            & 0.6944 & 0.6929 & 0.6950 & 0.6982 & 0.6964 & 0.6955 \\
        \midrule
        Edge rewiring
            & G-NAC
            & 0.7951 & 0.6684 & 0.5729 & 0.5306 & 0.4875 & 0.4240 \\
        & Shared spectral
            & 0.6944 & 0.5914 & 0.5486 & 0.5063 & 0.4479 & 0.3583 \\
        \bottomrule
    \end{tabular}%
    }
\end{table*}

\subsubsection{Measurement Noise}

\begin{figure}[htbp]
    \centering
    \includegraphics[width=0.85\linewidth]{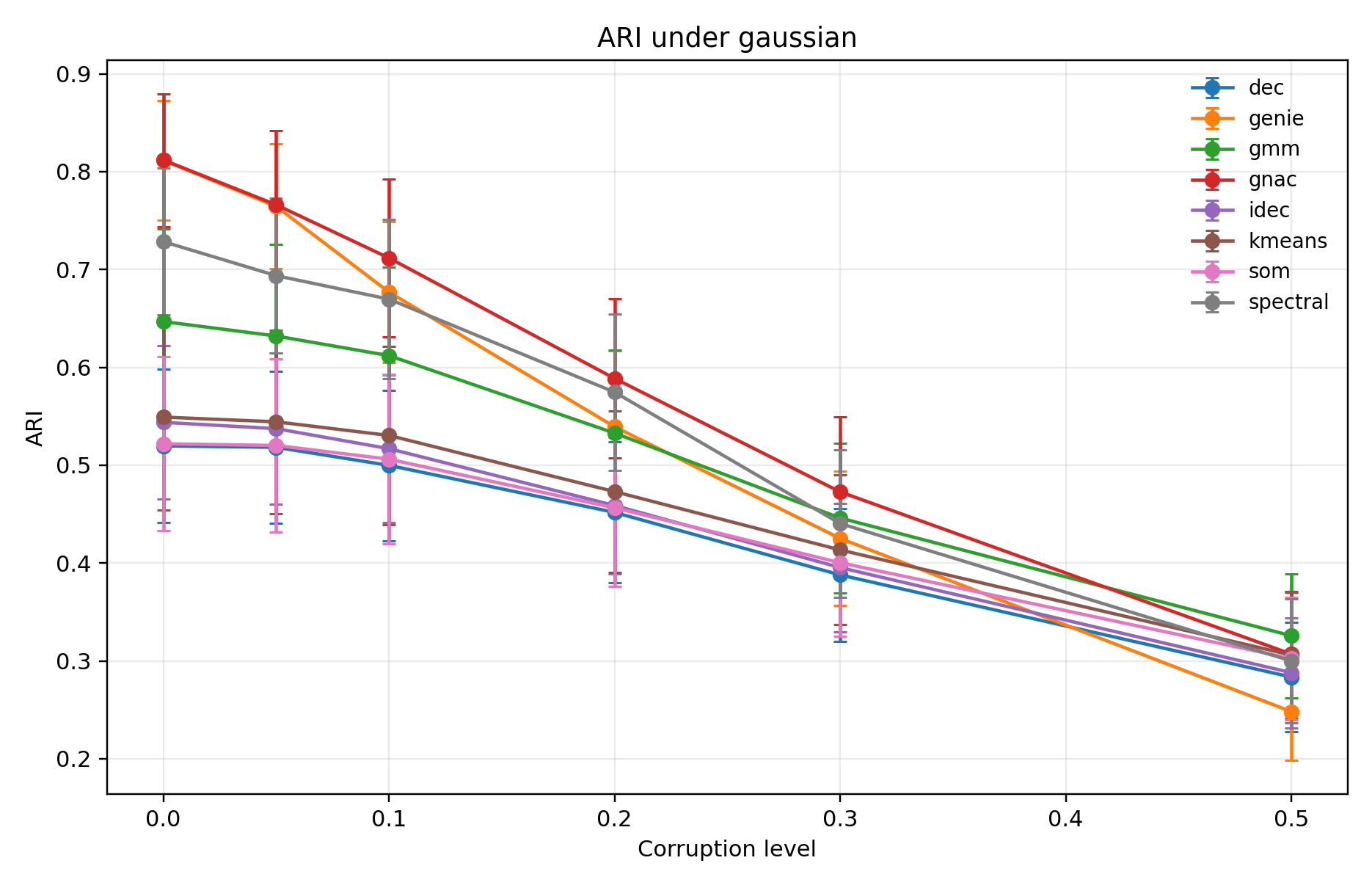}
    \caption{Mean ARI as noise increases.}
    \label{fig:noise-curves}
\end{figure}

G-NAC mean ARI decreased progressively from $0.7951$ on clean data to
$0.7571$, $0.7090$, $0.5899$, $0.4786$, and $0.3119$ as corruption
increased from $r=0.05$ to $0.50$. Every nonzero condition differed
significantly from clean performance after Holm correction
($p_{\mathrm{Holm}}\leq0.003$); Figure~\ref{fig:noise-curves}
shows the corresponding curves.

Relative to Genie, G-NAC's dataset-level mean ARI advantage was
$0.0494$, $0.0476$, and $0.0588$ at $r=0.20$, $0.30$, and $0.50$,
respectively, with $p_{\mathrm{Holm}}=0.0188$, $0.0031$, and
$3.6\times10^{-5}$. This does not imply uniformly slower degradation
than every baseline: at $r=0.50$, for example, GMM obtained mean ARI
$0.3353$ versus $0.3119$ for G-NAC.

\subsubsection{Missing Versus Incorrect Graph Relationships}

Edge deletion produced little aggregate degradation. Mean G-NAC ARI
remained within $0.004$ of clean performance and reached $0.7990$ after
deletion of 50\% of directed neighbor slots; none of the five nonzero
levels differed significantly from clean after Holm correction. Mean
component count changed only from $1.96$ to $1.99$, with no isolated
nodes. The shared-support spectral control likewise remained near ARI
$0.69$. G-NAC maintained a mean advantage of approximately
$0.086$--$0.090$ over this control at every level
($p_{\mathrm{Holm}}\leq0.0019$), but their degradation rates were not
significantly different.

Rewiring was substantially more damaging. G-NAC mean ARI fell from
$0.7951$ to $0.6684$ at 5\% rewiring and $0.4240$ at 50\%, with every
nonzero condition differing from clean
($p_{\mathrm{Holm}}\leq3.4\times10^{-5}$). Mean component count also
fell from $1.96$ to $1.00$ at 5\%, eliminating component-resolved
readouts. G-NAC remained above the shared-support spectral control at
every level, although only the 5\% comparison remained significant
after family-wise correction.

Under this corruption-and-retraining protocol, deletion of directed
neighbor slots was therefore tolerated substantially better than their
replacement by incorrect relationships. This asymmetry indicates
dependence on trustworthy local topology rather than an ability to
recover useful structure from arbitrary connectivity.

\subsubsection{Component-Resolved Readout Sensitivity}

The component-resolved branch was activated on 15 of the 73 clean
tasks, all with ARI $=1$. Because these outputs are determined by the
fixed graph connectivity, the analysis was repeated without them,
leaving 58 tasks from 47 datasets. Mean task-level ARI was $0.7421$
for G-NAC, $0.7623$ for Genie, $0.6981$ for GMM, and $0.6741$ for
spectral clustering.

G-NAC remained significantly different from spectral clustering at
the dataset level ($\Delta$ARI $=0.0619$, 95\% CI
$[0.0178,0.1101]$, $p_{\mathrm{Holm}}=0.0392$), whereas comparisons
with Genie ($-0.0206$, $p_{\mathrm{Holm}}=0.573$) and GMM
($0.0610$, 95\% CI $[-0.0140,0.1465]$,
$p_{\mathrm{Holm}}=0.455$) were not significant.

Rewiring sensitivity also persisted: mean G-NAC ARI followed
$0.7421\rightarrow0.6663\rightarrow0.6182\rightarrow0.5887
\rightarrow0.5413\rightarrow0.4645$ from $r=0$ to $0.50$.
Thus, the component-resolved cases contribute to aggregate performance
but do not fully explain the spectral comparison or sensitivity to
incorrect graph relationships.

\subsection{Time and Memory Scalability}
\label{sec:scalability_experiment}

A third experiment isolated scaling with the number of graph cells
$N$. A single balanced isotropic Gaussian-mixture distribution with
$K=8$ clusters and $d=16$ observed features was sampled once at
$N_{\max}=100{,}000$. Cluster centers were placed at radius $4.0$ and
each component used unit isotropic standard deviation. Smaller datasets
were nested prefixes of the same master sample, so
\begin{equation}
    X_{1k}\subset X_{2k}\subset X_{5k}\subset\cdots\subset X_{100k}.
\end{equation}
The tested sizes were
\begin{equation}
    N\in\{1{,}000,2{,}000,5{,}000,10{,}000,20{,}000,50{,}000,100{,}000\}.
\end{equation}
At every scale G-NAC retained the frozen G-NAC configuration used throughout this study:
$k=20$, 16 recurrent training steps, 100 epochs, FP32 arithmetic,
activation checkpointing, and the same adaptive inference rule.
Three algorithm seeds ($7$, $17$, and $27$) were evaluated per size.

The experiment separately timed neighborhood-graph construction,
training after graph construction, adaptive inference, and final
readout. CUDA peak allocated and reserved memory were measured with
PyTorch, while process resident-set size was sampled independently.
CUDA operations were synchronized around timed stages. All runs were
performed on an NVIDIA GeForce RTX 5070 with 12~GB device memory under
Windows 11. No component-resolved readout occurred at any scale, so the
large-$N$ timing curve reflects the nontrivial spectral-readout path.
Table~\ref{tab:scalability} summarizes the resulting timing and GPU-memory
measurements.

\begin{table*}[t]
    \centering
    \caption{Empirical G-NAC scalability. Times and memory values are medians
    across three algorithm seeds. Power exponents $\alpha$ are fitted over
    $N\geq5{,}000$ using
    $\log y=\alpha\log N+\beta$ to reduce the influence of fixed startup
    overhead at the two smallest sizes. CUDA memory is reported in MiB.}
    \label{tab:scalability}
    \small
    \resizebox{\textwidth}{!}{%
    \begin{tabular}{lrrrrrrr}
        \toprule
        Quantity & 5k & 10k & 20k & 50k & 100k & $\alpha$ & $R^2$ \\
        \midrule
        Total fit time (s)
            & 15.27 & 30.75 & 60.55 & 153.27 & 333.34 & 1.023 & 0.9995 \\
        Graph construction (s)
            & 0.54 & 1.11 & 2.48 & 7.51 & 17.25 & 1.166 & 0.9993 \\
        Fit excluding graph (s)
            & 14.74 & 29.64 & 58.05 & 145.82 & 315.93 & 1.016 & 0.9995 \\
        End-to-end time (s)
            & 15.80 & 31.88 & 62.86 & 159.55 & 346.79 & 1.025 & 0.9995 \\
        Peak CUDA allocated (MiB)
            & 364.8 & 717.7 & 1421.9 & 3548.5 & 7085.3 & 0.991 & 1.0000 \\
        Peak CUDA reserved (MiB)
            & 482 & 942 & 1930 & 4732 & 9230 & 0.989 & 0.9999 \\
        \bottomrule
    \end{tabular}%
    }
\end{table*}

For $N\geq5{,}000$, total fit time scaled empirically as $N^{1.023}$,
fit time excluding graph construction as $N^{1.016}$, and end-to-end
time as $N^{1.025}$. Graph construction was mildly superlinear
($N^{1.166}$) but required only $17.25$~s of the $333.34$~s median
fit time at $N=100{,}000$. Peak CUDA allocated and reserved memory
scaled as $N^{0.991}$ and $N^{0.989}$, reaching $7085$~MiB
($6.92$~GiB) and $9230$~MiB ($9.01$~GiB), respectively. Median
end-to-end execution at this scale was $346.79$~s.

Mean ARI remained between $0.961$ and $0.974$ across all seven sizes
and reached $0.9736$ at $N=100{,}000$. Near-linear scaling also
persisted when the static edge-difference cache was disabled at
$100{,}000$ nodes after exceeding its configured 128~MB cap. These
measurements are hardware- and implementation-specific, but indicate
approximately linear empirical time and GPU-memory scaling over the
tested range for fixed graph degree and model configuration.

\subsection{Sparse-to-Full Inductive Rule Transfer}
\label{sec:sparse_transfer}

The final experiment tested whether a cellular rule must be trained at
its eventual deployment scale. Rules learned on source graphs of
varying size were frozen and deployed without optimization on
independent 100,000-node target graphs from the same generating process.

\subsubsection{Experimental Design}

Two $d=16$, $K=8$ synthetic families were used: an isotropic Gaussian
mixture and a more difficult noisy-blobs family. Ten independent
source--target pairs were generated per family with matched generating
parameters and disjoint observations. Targets contained
$N_{\mathrm{target}}=100{,}000$ observations, while nested source
samples used
\[
N_{\mathrm{source}}
\in
\{1{,}000,2{,}000,5{,}000,10{,}000,20{,}000,50{,}000,100{,}000\}.
\]
Each source graph was constructed only from observations available at
that size. Three training seeds were evaluated using the same fixed
G-NAC configuration as the preceding experiments.

After training, only the learned transition-rule parameters were
transferred. Fresh cellular states were initialized for deployment
either on the complete source realization or on the independent
100,000-node target. Direct target-trained G-NAC, target rank-affinity
spectral clustering, $K$-means, and an untrained randomly initialized
G-NAC rule served as controls; $K=8$ was used only by the final
readout.

Statistical comparisons used the independent source--target pair as
the experimental unit after averaging its three training seeds. An ARI
difference of $0.03$ was selected prospectively as the practical
noninferiority margin relative to complete 100,000-node source
training.

\subsubsection{Transfer across graph size and realization}

Table~\ref{tab:sparse_transfer} summarizes independent-target
performance as a function of source training size. On the Gaussian
mixture, target performance was nearly invariant to source size. A rule
trained using only 1,000 observations achieved mean ARI $0.9729$ after
deployment to an independent 100,000-node target, compared with
$0.9739$ for rules trained on all 100,000 source observations.

The more difficult noisy-blobs family exhibited a clearer sample-size
dependence. Rules trained on 1,000 and 2,000 source observations achieved
mean independent-target ARI values of $0.7343$ and $0.7240$,
respectively, corresponding to reductions of $0.0366$ and $0.0469$
relative to complete-source training. Both exceeded the prespecified
$0.03$-ARI practical tolerance. In contrast, increasing the source
sample to 5,000 observations raised mean target ARI to $0.7621$,
compared with $0.7709$ for complete 100,000-node source training. The
paired difference was

\[
\Delta_{\mathrm{sparse}}
=
-0.0089,
\qquad
95\%~\mathrm{CI}
=
[-0.0114,-0.0064],
\]

placing the 5,000-node condition comfortably within the prespecified
practical noninferiority margin. The remaining difference continued to
decrease with source size, reaching $-0.0025$ at 20,000 observations
and approximately zero at 50,000 observations.

\begin{table}[t]
\centering
\caption{Sparse-to-full transfer to independent 100,000-node target
graphs. $\Delta_{\mathrm{noisy}}$ denotes the mean ARI difference
relative to a rule trained on the complete 100,000-node source graph.
Training time is shown for the noisy-blobs source condition.}
\label{tab:sparse_transfer}
\begin{tabular}{rrrrr}
\toprule
$N_{\mathrm{source}}$
& Gaussian ARI
& Noisy ARI
& $\Delta_{\mathrm{noisy}}$
& Fit time (s) \\
\midrule
1,000   & 0.9729 & 0.7343 & -0.0366 & 10.0 \\
5,000   & 0.9735 & 0.7621 & -0.0089 & 15.2 \\
20,000  & 0.9737 & 0.7685 & -0.0025 & 60.0 \\
100,000 & 0.9739 & 0.7709 &  0.0000 & 328.5 \\
\bottomrule
\end{tabular}
\end{table}

Importantly, the transfer behavior was not explained by reuse of source
observations. Across source sizes, performance on the overlapping
within-realization graph and the completely disjoint target graph was
nearly identical. For example, at $N_{\mathrm{source}}=5{,}000$, the
mean noisy-blobs ARI was $0.7647$ under within-realization deployment
and $0.7621$ on the independent target. Similar differences of only a
few thousandths of ARI were observed throughout the source-size sweep.

Full-scale transfer further isolated the cost of changing graph
realizations from that of sparse training. For the Gaussian family, a
rule trained on an independent 100,000-node source graph differed from
a rule retrained directly on the target by only

\[
\Delta_{\mathrm{transfer}}=-0.00007
\]

ARI, with a 95\% confidence interval of approximately
$[-0.00015,0.00000]$. For noisy blobs, the corresponding difference was

\[
\Delta_{\mathrm{transfer}}=+0.00027,
\]

with a 95\% confidence interval of
$[-0.00112,0.00219]$. Thus, once the source graph was sufficiently
sampled, transferring the learned cellular rule to an unseen graph
incurred essentially no additional performance penalty relative to
retraining G-NAC directly on that graph.

\subsubsection{Sparse Rule Learning and Computational Cost}

On noisy blobs, 5,000-node source training required approximately
15.2~s and 365~MiB peak allocated CUDA memory, compared with
328.5~s and 7.0~GiB at 100,000 nodes. The 5\% source condition
therefore reduced training time by approximately $21.6\times$ and
memory by roughly $19$--$20\times$ while reducing independent-target
ARI by less than $0.009$. No component-resolved source readouts
occurred.

Sparse-source self-performance could be substantially lower than
target performance of the same rule. At 1,000 nodes, noisy-blobs
self-ARI was $0.5506$, whereas independent-target ARI was $0.7343$;
the difference approached zero by 100,000 observations. Thus, rule
quality and the quality of the finite graph on which that rule is
expressed appear partially separable.

\subsubsection{Relation to Local Graph Geometry}

Local normalization provides one possible explanation for scale
transfer. On noisy blobs, mean raw edge distance decreased from
approximately $5.83$ at 1,000 source observations to $4.00$ on the
100,000-node target, whereas the normalized distance
\[
\widetilde d_{ij}
=
\frac{d_{ij}}
{\frac{1}{2}(s_i+s_j)+\epsilon}
\]
remained approximately $1.0485$ and $1.0487$. The Gaussian family
showed similar behavior. This association is consistent with reduced
dependence on sampling density but does not establish a causal role for
normalization.

Other graph properties changed with density. On noisy blobs,
cross-cluster edges decreased from approximately $0.386$ to $0.189$,
while target-domain boundary AUC increased from approximately $0.634$
after 1,000-node source training to $0.749$ after complete-source
training. Improved local neighborhood fidelity may therefore contribute
to the observed source-size dependence.

The controls constrain the interpretation of the transfer result.
On 100,000-node targets, direct rank-affinity spectral clustering
achieved ARI approximately $0.9745$ and $0.7670$ on Gaussian and noisy
data, respectively; $K$-means achieved $0.9768$ and $0.6622$; an
untrained G-NAC rule achieved $0.9728$ and $0.7634$; and transferred
complete-source G-NAC achieved $0.9739$ and $0.7709$. Thus, these
experiments establish transferability and sample efficiency of fitted
G-NAC rules, not the necessity of fitting for strong performance on
these synthetic distributions.

\subsection{Summary of Experimental Findings}
\label{sec:experimental_summary}

Across the 73 benchmark tasks, G-NAC achieved the highest overall mean
and median ARI among the eleven evaluated methods using a single fixed
hyperparameter configuration. Dataset-level paired analysis showed
significant differences from nine baselines, including ordinary
spectral clustering, while no significant difference was observed
between G-NAC and Genie. These results establish the complete G-NAC
pipeline as competitive across the evaluated benchmark, although they
do not isolate the contribution of recurrent learning from the graph
construction and rank-based readout.

The subsequent experiments clarify the conditions and costs associated
with this performance. G-NAC was comparatively insensitive to deletion
of directed neighbor slots under the tested retraining protocol, but
degraded progressively under erroneous rewiring and measurement noise,
indicating greater sensitivity to incorrect than incomplete local
relationships. For fixed graph degree and model configuration, training
time and GPU memory scaled approximately linearly over
$5{,}000$--$100{,}000$ nodes. Finally, learned transition rules
transferred without retraining to substantially larger, independently
sampled graphs from the same generating processes. On the more difficult
noisy-blobs family, a rule trained on 5,000 nodes achieved target
performance within $0.009$ ARI of 100,000-node source training while
requiring approximately $21.6\times$ less training time and roughly
$19$--$20\times$ less peak allocated GPU memory.

Taken together, the experiments characterize G-NAC as a competitive but
computationally more expensive clustering procedure whose performance
depends on meaningful local graph structure, whose empirical cost scales
approximately linearly over the tested range, and whose learned cellular
rule can transfer across graph size and independent graph realizations
under matched generating conditions.

\section{Discussion}

G-NAC is best understood as a learned dynamical process operating over
local relational structure rather than simply as a neural embedding
method. A neighborhood graph defines local interactions, while repeated
application of a shared graph-neural cellular rule reorganizes node
states into a cluster-revealing relational representation. Clusters
therefore emerge from collective graph-state evolution rather than
direct prediction of assignments.

\subsection{Clustering as Learned Domain Formation}

A useful interpretation of G-NAC is that clusters correspond to domains
of an evolving cellular system. The graph defines the interaction
topology, while the learned transition rule determines how information
propagates across it. The unsupervised objective encourages local
coherence while penalizing representational collapse, allowing distinct
regions of the graph to emerge through the recurrent dynamics.

The partition is consequently a readout of these dynamics rather than
their direct optimization target. This perspective also explains why
the useful information need not reside in a globally stationary latent
coordinate system: cluster structure can instead be expressed through
relationships between cells.

\subsection{Relational Rank Stability Rather Than Coordinate Convergence}

Successful G-NAC inference does not necessarily correspond to a
fixed-point attractor. In several experiments, domain coordinates
continued to evolve after their cluster-relevant relational ordering
changed much more slowly. Coordinate-wise convergence is therefore not
required by the present inference procedure.

This behavior motivates both the empirical rank-stability criterion and
the final edge-rank affinity. The former detects high rank correlation
between successive edge-distance configurations, while the latter
constructs the partitioning affinity from their ordinal relationships.
The threshold $\rho\geq0.95$ is an empirical stopping rule rather than
a convergence certificate or guarantee of future partition stability.

At fixed edge support and tie convention, the rank affinity is invariant
to strictly increasing transformations of edge distances. Changes in
latent scale therefore do not affect the readout when edge ordering is
preserved, providing a readout compatible with relational rather than
coordinate convergence.

\subsection{The Role of the Initial Graph}

G-NAC cannot recover arbitrary structure absent from its interaction
graph. Sparse graphs may preserve locally correct relationships while
fragmenting manifolds, whereas denser graphs can restore connectivity
while introducing cross-cluster shortcuts. Consequently, strong local
edge discrimination does not imply that the fixed graph contains
sufficient global information to recover the desired partition.

G-NAC presently transforms states and learned relations on a fixed
topology rather than rewiring that topology. Its natural operating
regime is therefore a graph with meaningful local connectivity but
sufficient ambiguity for relational refinement to be useful. In this
sense, G-NAC is better viewed as learned graph re-embedding or
refinement than as a general solution to graph construction.

\subsection{Transferability of the Cellular Rule}

Sparse-to-full transfer suggests that the learned dynamics are not
intrinsically tied to the graph realization used for optimization.
Under matched synthetic generating conditions, sufficiently sampled
source graphs produced frozen rules that transferred to substantially
larger independent graphs with little additional loss relative to
target-scale training.

This behavior is consistent with learning a shared local rule rather
than node-specific parameters. Locally normalized geometric inputs also
changed substantially less across sampling densities than raw neighbor
distances, providing one plausible mechanism for scale transfer.
Moreover, some sparse-source rules performed substantially better after
deployment to densely sampled targets than on their own source graphs,
suggesting partial separation between the quality of a learned rule and
the finite graph substrate on which it is expressed. Whether these
properties extend beyond matched generating processes remains open.

\subsection{Robustness and Stability}

The corruption experiments distinguish incomplete from incorrect local
topology. Deletion of up to 50\% of directed neighbor slots before union
symmetrization produced no statistically detectable aggregate loss
under retraining, whereas rewiring those slots to non-neighbors caused
substantial degradation. Because the shared-support spectral control
was also insensitive to deletion, this robustness is better attributed
to redundancy of the neighborhood substrate than to unique error
correction by the recurrent dynamics.

These experiments characterize robustness of the fitting procedure,
since G-NAC was retrained after each corruption. They indicate that the
method tolerates substantial loss of local relationships when those
remaining are meaningful but is sensitive to false connectivity.
Gaussian feature noise provides an intermediate case because it
simultaneously perturbs observations and their induced graph geometry.

G-NAC also showed relatively low stochastic variability across the
benchmark. Difficult cases nevertheless produced distinct outcomes,
consistent with the possibility of multiple dynamical regimes under
different initial conditions. Characterizing these regimes more
formally remains future work.

\subsection{Computational Considerations}

G-NAC is substantially more expensive than conventional clustering
because its recurrent transition rule must be optimized before
partitioning. Its computational motivation therefore lies in problems
where relational structure justifies this additional cost rather than
in replacing inexpensive methods on simple cluster geometries.

For fixed graph degree and model configuration, however, the controlled
experiment showed approximately linear empirical scaling in training
time and GPU memory over $5{,}000$--$100{,}000$ nodes. Graph
construction and spectral readout remain separate costs whose scaling
can become important beyond this range.

Sparse rule learning offers a complementary reduction in training cost.
On noisy blobs, 5,000-node source training retained independent-target
ARI within $0.009$ of 100,000-node source training while requiring
approximately $21.6\times$ less training time and $19$--$20\times$
less peak allocated GPU memory. This does not reduce deployment-graph
construction, recurrent inference, or final readout costs, but suggests
that gradient-based rule learning can be decoupled from deployment
scale under suitable conditions.

\subsection{Limitations and Future Directions}

The experiments evaluate the complete G-NAC pipeline more directly than
they isolate its individual mechanisms. Ordinary spectral clustering
does not simultaneously control graph support, affinity construction,
and representation, while graph-only rank affinities and untrained
recurrent controls were already strong on the synthetic transfer
families. The benchmark therefore establishes performance of the
complete procedure rather than attributing its differences specifically
to recurrence, learned conductance, or individual objective terms.
Matched mechanism ablations remain an important next step.

G-NAC also depends fundamentally on initial graph quality. Missing
relationships and cross-cluster shortcuts can constrain what fixed-
topology dynamics can recover, motivating future work on jointly learned
or adaptive topology. In addition, $K$ is currently required by the
final readout despite being absent from training and recurrent
inference; estimating the number of emergent domains directly from the
evolved relational state would provide a fully unsupervised readout.

The demonstrated rule transfer is limited to independent graphs sharing
the same synthetic generating parameters. It establishes inductive
graph transfer within a generating process, not transfer across
distribution shift, unrelated datasets, or application domains.
Testing reuse across changes in feature distributions and graph
geometry is therefore necessary.

Finally, the current formulation uses a single graph scale and level of
cellular organization. Hierarchical graph-neural cellular systems, in
which coarse and fine domains interact, provide a natural extension for
multiscale structured data.

\section{Conclusion}

We introduced G-NAC as an unsupervised clustering formulation in which a
shared graph-neural cellular transition rule reorganizes observations
before a final partition is read from learned graph-edge relationships.
The method treats clustering as emergent domain formation on a fixed
neighborhood graph rather than direct prediction of cluster labels. On
the 73-task benchmark, the complete pipeline was competitive with the
strongest evaluated baseline, while corruption, scalability, and
transfer experiments characterized the graph-quality assumptions and
computational conditions under which the approach remains useful.

The results also delimit the present claims. G-NAC depends on meaningful
local graph structure, requires $K$ at the final readout, and is more
computationally expensive than conventional clustering. Strong graph-only
and untrained controls on the synthetic transfer tasks, together with
component-resolved benchmark cases, mean that the observed performance
cannot be attributed uniformly to learned recurrence alone. The next
steps are therefore matched mechanism ablations, adaptive or learned
graph construction, automatic estimation of emergent domain count, and
tests of cellular-rule transfer beyond matched generating processes.

\section*{Code and Data Availability}
A public implementation of G-NAC and the experimental scripts supporting
this study are being prepared for release. Until that release, materials
needed to reproduce the reported experiments are available from the authors
upon reasonable request. The primary benchmark datasets are drawn from the
publicly available, versioned Clustering Benchmarks suite cited in the text.

\bibliographystyle{plainnat}
\bibliography{references}

\end{document}